\documentclass[11pt,a4paper]{article}
\usepackage[hyperref]{naaclhlt2018}
\usepackage{times}
\usepackage{latexsym}
\usepackage{graphicx}
\usepackage{url}
\usepackage{amsmath}
\usepackage{amssymb}
\usepackage[linesnumbered]{algorithm2e}
\usepackage{color}
\usepackage{tabularx}
\usepackage{tablefootnote}

\aclfinalcopy 

\title{A Graph-to-Sequence Model for AMR-to-Text Generation}

\author{Linfeng Song$^1$, Yue Zhang$^3$, Zhiguo Wang$^2$ \and Daniel Gildea$^1$ \\
  $^1$Department of Computer Science, University of Rochester, Rochester, NY 14627 \\
  $^2$IBM T.J. Watson Research Center, Yorktown Heights, NY 10598 \\
  $^3$Singapore University of Technology and Design}

\date{}

\begin{document}
\maketitle
\begin{abstract}
  The problem of AMR-to-text generation is to recover a text representing the same meaning as an input AMR graph.
  The current state-of-the-art method uses a sequence-to-sequence model, leveraging LSTM for encoding a linearized AMR structure. 
  Although it is able to model non-local semantic information, a sequence LSTM can lose information from the AMR graph structure, and thus faces challenges with large graphs, which result in long sequences. 
  We introduce a neural graph-to-sequence model, using a novel LSTM structure for directly encoding graph-level semantics.
  On a standard benchmark, our model shows superior results to existing methods in the literature.
\end{abstract}

\section{Introduction}

\begin{figure}
\centering
\includegraphics[scale=0.6]{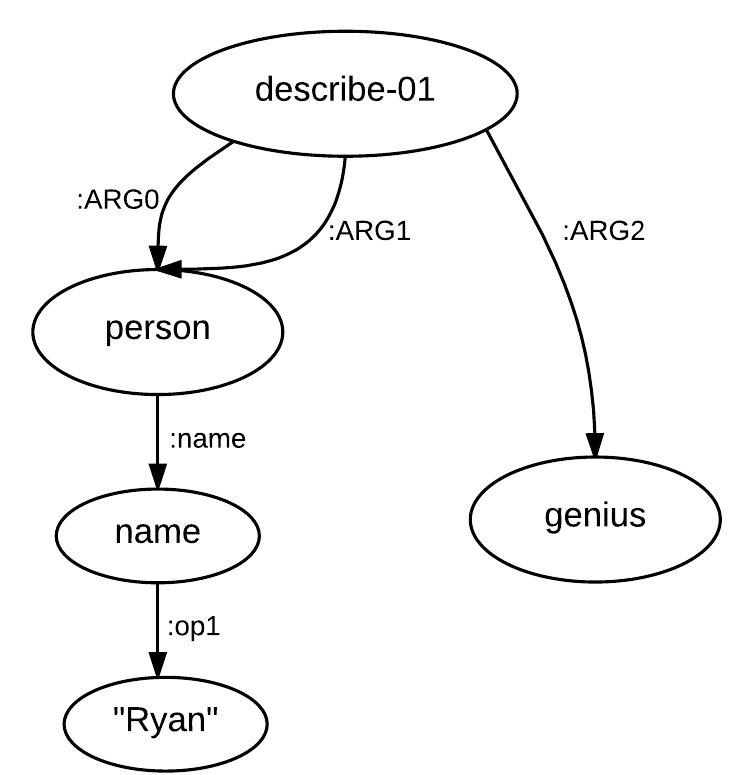}
\caption{An example of AMR graph meaning ``Ryan's description of himself: a genius.''}
\label{fig:example_amr}
\end{figure}

Abstract Meaning Representation (AMR) \cite{banarescu-EtAl:2013:LAW7-ID} is a semantic formalism that encodes the meaning of a sentence as a rooted, directed graph.
Figure \ref{fig:example_amr} shows an AMR graph in which the nodes (such as ``describe-01'' and ``person'') represent the concepts, and edges (such as ``:ARG0'' and ``:name'') represent the relations between concepts they connect.
AMR has been proven helpful on other NLP tasks, such as machine translation \cite{jones2012semantics,tamchyna-quirk-galley:2015:S2MT}, question answering \cite{mitra2015addressing}, summarization \cite{takase-EtAl:2016:EMNLP2016} and event detection \cite{li-EtAl:2015:CNewsStory}.

The task of AMR-to-text generation is to produce a text with the same meaning as a given input AMR graph. 
The task is challenging as word tenses and function words are abstracted away when constructing AMR graphs from texts.
The translation from AMR nodes to text phrases can be far from literal.
For example, shown in Figure \ref{fig:example_amr}, ``Ryan'' is represented as ``(p / person :name (n / name :op1 ``Ryan''))'', and ``description of'' is represented as ``(d / describe-01 :ARG1 )''.

While initial work used statistical approaches \cite{jeff2016amrgen,pourdamghani-knight-hermjakob:2016:INLG,song-EtAl:2017:Short,lampouras-vlachos:2017:SemEval,mille-EtAl:2017:SemEval,gruzitis-gosko-barzdins:2017:SemEval}, recent research has demonstrated the success of deep learning, and in particular the sequence-to-sequence model \cite{sutskever2014sequence}, which has achieved the state-of-the-art results on AMR-to-text generation \cite{konstas-EtAl:2017:Long}. 
One limitation of sequence-to-sequence models,
however, is that they require serialization of input AMR graphs, which adds to the challenge of representing graph structure information, especially when the graph is large.
In particular, closely-related nodes, such as parents, children and siblings can be far away after serialization.
It can be difficult for a linear recurrent neural network to automatically induce their original connections from bracketed string forms.

To address this issue, we introduce a novel graph-to-sequence model, where a graph-state LSTM is used to encode AMR structures directly.
To capture non-local information, the encoder performs graph state transition by information exchange between connected nodes, with a graph state consisting of all node states.
Multiple recurrent transition steps are taken so that information can propagate non-locally, and LSTM \cite{hochreiter1997long} is used to avoid gradient diminishing and bursting in the recurrent process.
The decoder is an attention-based LSTM model with a 
copy mechanism \cite{gu-EtAl:2016:P16-1,gulcehre-EtAl:2016:P16-1}, which helps copy sparse tokens (such as numbers and named entities) from the input.

Trained on a standard dataset (LDC2015E86), our model surpasses a strong sequence-to-sequence baseline by 2.3 BLEU points, demonstrating the advantage of graph-to-sequence models for AMR-to-text generation compared to sequence-to-sequence models.
Our final model achieves a BLEU score of 23.3 on the test set, which is 1.3 points higher than the  existing state of the art \cite{konstas-EtAl:2017:Long} trained on the same dataset.
When using gigaword sentences as additional training data, our model is consistently better than \newcite{konstas-EtAl:2017:Long} using the same amount of gigaword data, showing the effectiveness of our model on large-scale training set.

We release our code and models at \url{https://github.com/freesunshine0316/neural-graph-to-seq-mp}.

\section{Baseline: a seq-to-seq model}
\label{sec:base}

Our baseline is a sequence-to-sequence model, which follows the encoder-decoder framework of \newcite{konstas-EtAl:2017:Long}.

\subsection{Input representation}
\label{sec:base_inp}

Given an AMR graph $G=(V,E)$, where $V$ and $E$ denote the sets of nodes and edges, respectively, we use the depth-first traversal of \newcite{konstas-EtAl:2017:Long} to linearize it to obtain a sequence of tokens $v_1, \dots, v_N$, where $N$ is the number of tokens.
For example, the AMR graph in Figure 1 is serialized as ``describe :arg0 ( person :name ( name :op1 ryan )  )  :arg1 person :arg2 genius''.
We can see that the distance between ``describe'' and ``genius'', which are directly connected in the original AMR,  becomes 14 in the serialization result.

A simple way to calculate the representation for each token $v_j$ is using its word embedding $e_j$:
\begin{equation}
x_j = W_1 e_{j} + b_1 \textrm{,}
\label{eq:base_inp}
\end{equation}
where $W_1$ and $b_1$ are model parameters for compressing the input vector size.

To alleviate the data sparsity problem and obtain better word representation as the input, we also adopt a forward LSTM over the characters of the token, and concatenate the last hidden state $h_{j}^c$ with the word embedding:
\begin{equation}
x_j = W_1 \Big( [e_{j}; h_{j}^c] \Big) + b_1
\label{eq:base_inp_2}
\end{equation}

\subsection{Encoder}
\label{sec:base_enc}

The encoder is a bi-directional LSTM applied on the linearized graph by depth-first traversal, as in \newcite{konstas-EtAl:2017:Long}.
At each step $j$, the current states $\overleftarrow{h_j}$ and $\overrightarrow{h_j}$ are generated given the previous states $\overleftarrow{h_{j+1}}$ and $\overrightarrow{h_{j-1}}$ and the current input $x_j$:
\begin{align*}
\overleftarrow{h_j} &= \textrm{LSTM}(\overleftarrow{h_{j+1}}, x_j) \\
\overrightarrow{h_j} &= \textrm{LSTM}(\overrightarrow{h_{j-1}}, x_j)
\end{align*}

\subsection{Decoder}
\label{sec:base_dec}

We use an attention-based LSTM decoder \cite{bahdanau2015neural}, where the attention memory ($A$) is the concatenation of the attention vectors among all input words. 
Each attention vector $a_j$ is the concatenation of the encoder states of an
input token in both directions ($\overleftarrow{h_j}$ and $\overrightarrow{h_j}$) and its input vector ($x_j$):
\begin{align}
a_j &= [\overleftarrow{h_j}; \overrightarrow{h_j}; x_j] \\
A &= [a_1; a_2; \dots; a_N]
\end{align}
where $N$ is the number of input tokens.

The decoder yields an output sequence $w_1, w_2, \dots, w_M$ by calculating a sequence of hidden states $s_1, s_2 \dots, s_M$ recurrently.
While generating the $t$-th word, the decoder considers five factors: 
(1) the attention memory $A$; 
(2) the previous hidden state of the LSTM model $s_{t-1}$; 
(3) the embedding of the current input (previously generated word) $e_{t}$; 
(4) the previous context vector $\mu_{t-1}$, which is calculated with attention from $A$; 
and (5) the previous coverage vector $\gamma_{t-1}$, which is the accumulation of all attention distributions so far \cite{tu-EtAl:2016:P16-1}. 
When $t=1$, we initialize $\mu_{0}$ and $\gamma_{0}$ as zero vectors, set $e_{1}$ to the embedding of the start token ``$<$s$>$'', and  $s_{0}$ as the average of all encoder states.

For each time-step $t$, the decoder feeds the concatenation of the embedding of the current input $e_{t}$ and the previous context vector $\mu_{t-1}$ into the LSTM model to update its hidden state.
Then the attention probability $\alpha_{t,i}$ on the attention vector $a_i \in A$ for the time-step is calculated as:
\begin{align*}
\epsilon_{t,i} &= v_2^T \tanh(W_a a_i + W_s s_t + W_{\gamma} \gamma_{t-1} + b_2) \\
\alpha_{t,i} &= \frac{\exp(\epsilon_{t,i})}{\sum_{j=1}^N\exp(\epsilon_{t,j})} 
\end{align*}
where $W_a$, $W_s$, $W_{\gamma}$, $v_2$ and $b_2$ are model parameters.
The coverage vector $\gamma_t$ is updated by  $\gamma_t = \gamma_{t-1} + \alpha_t$, and the new context vector $\mu_t$ is calculated via $\mu_t = \sum_{i=1}^N \alpha_{t,i} a_{i}$.

The output probability distribution over a vocabulary at the current state is calculated by:
\begin{equation}
P_{vocab} = \textrm{softmax}(V_3[s_t,\mu_t]+b_3)\textrm{,}
\label{eq:pvocab}
\end{equation}
where $V_3$ and $b_3$ are learnable parameters, and the number of rows in $V_3$ represents the number of words in the vocabulary.

\section{The graph-to-sequence model}

Unlike the baseline sequence-to-sequence model,
we leverage a recurrent graph encoder to represent each input AMR, which directly models the graph structure without serialization. 

\subsection{The graph encoder}

\begin{figure}
\centering
\includegraphics[width=0.9\linewidth]{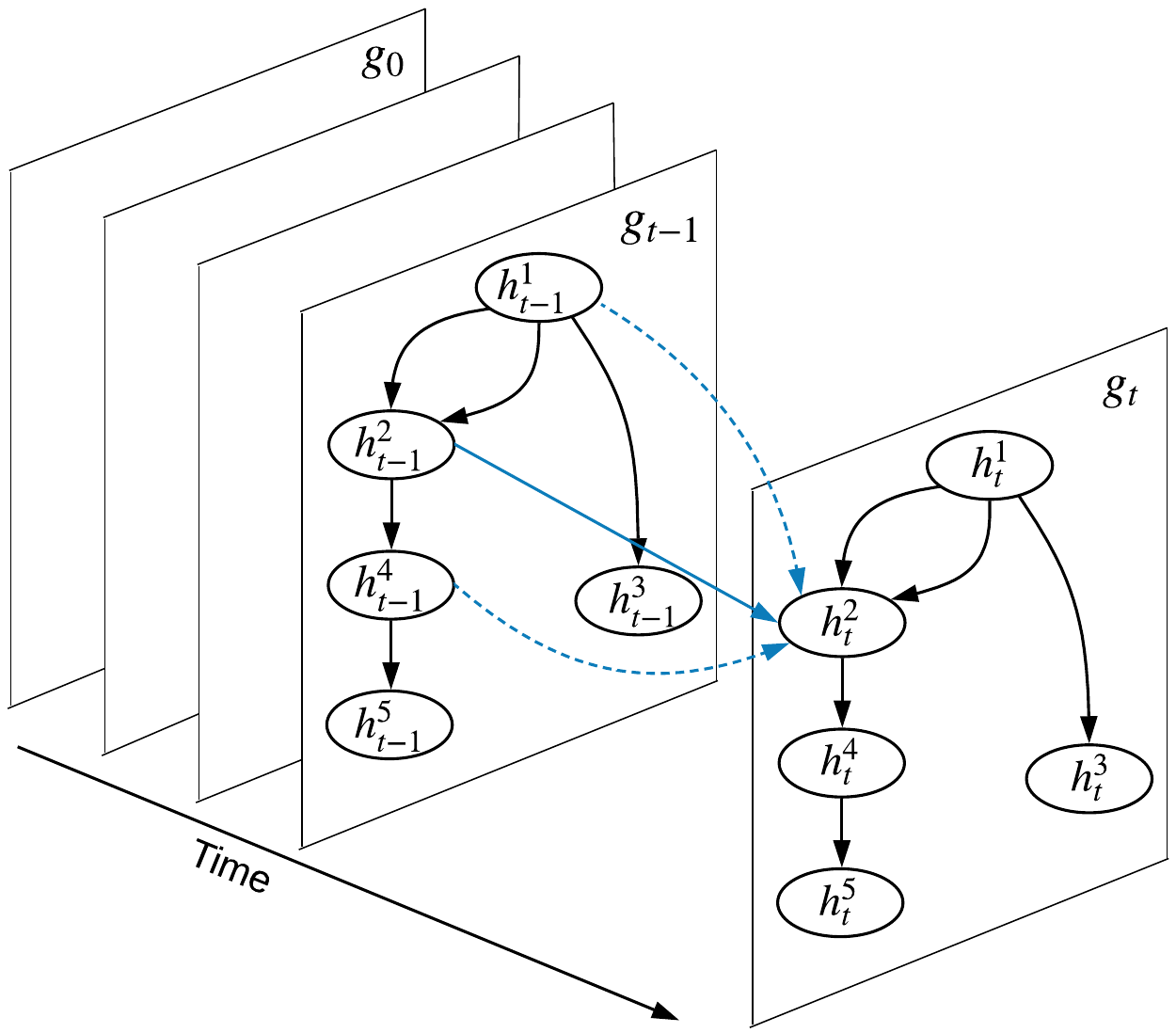}
\caption{Graph state LSTM.}
\label{fig:encoder}
\end{figure}

Figure \ref{fig:encoder} shows the overall structure of our graph encoder. 
Formally, given a graph $G=(V, E)$, we use a hidden state vector $h^j$ to represent each node $v_j \in V$. 
The state of the graph can thus be represented as:
\[
g = \{h^j\}|_{v_j \in V}
\]
In order to capture non-local interaction between nodes, we allow information exchange between nodes through a sequence of state transitions, leading to a sequence of states $g_0, g_1, \dots, g_t, \dots$, where $g_t = \{h_t^j\}|_{v_j \in V}$.
The initial state $g_0$ consists of a set of initial node states $h_0^j=h_0$, where $h_0$ is a hyperparameter of the model.

\subparagraph{State transition}
A recurrent neural network is used to model the state transition process. 
In particular, the transition from $g_{t-1}$ to $g_t$ consists of a
hidden state transition for each node, as shown in Figure \ref{fig:encoder}. 
At each state transition step $t$, we allow direct communication between a node and all nodes that are directly connected to the node. 
To avoid gradient diminishing or bursting, LSTM \cite{hochreiter1997long} is adopted, where a cell $c_t^j$ is taken to record memory for $h_t^j$. 
We use an input gate $i_t^j$, an output gate $o_t^j$ and a forget gate $f_t^j$ to control information flow from the inputs and to the output $h_t^j$.

The inputs include representations of edges that are connected to $v_j$, where $v_j$ can be either the source or the target of the edge.
We define each edge as a triple $(i,j,l)$, where $i$ and $j$ are indices of the source and target nodes, respectively, and $l$ is the edge label.
$x_{i,j}^l$ is the representation of edge $(i,j,l)$, detailed in Section \ref{sec:input}.
The inputs for $v_j$ are distinguished by incoming and outgoing edges, before being summed up:
\begin{equation*}
\begin{split}
x_j^{i} &= \sum_{(i,j,l)\in E_{in}(j)} x_{i,j}^l \\
x_j^{o} &= \sum_{(j,k,l)\in E_{out}(j)} x_{j,k}^l \textrm{,} \\
\end{split}
\end{equation*}
where $E_{in}(j)$ and $E_{out}(j)$ denote the sets of incoming and outgoing edges of $v_j$, respectively.

In addition to edge inputs, a cell also takes the hidden states of its incoming nodes and outgoing nodes during a state transition. 
In particular, the states of all incoming nodes and outgoing nodes are summed up before being passed to the cell and gate nodes:
\begin{equation*}
\begin{split}
h_j^{i} &= \sum_{(i,j,l)\in E_{in}(j)} h_{t-1}^{i} \\
h_j^{o} &= \sum_{(j,k,l)\in E_{out}(j)} h_{t-1}^{k} \textrm{,} \\
\end{split}
\end{equation*} 
Based on the above definitions of $x_j^{i}$, $x_j^{o}$, $h_j^{i}$ and $h_j^{o}$, the state transition from $g_{t-1}$ to $g_t$, as represented by $h_t^j$, can be defined as:
\begin{equation*}
\begin{split}
i_t^j &= \sigma(W_i x_j^{i} + \hat{W_i} x_j^{o} + U_i h_j^{i} + \hat{U_i} h_j^{o} + b_i) \textrm{,} \\
o_t^j &= \sigma(W_o x_j^{i} + \hat{W_o} x_j^{o} + U_o h_j^{i} + \hat{U_o} h_j^{o} + b_o) \textrm{,} \\
f_t^j &= \sigma(W_f x_j^{i} + \hat{W_f} x_j^{o} + U_f h_j^{i} + \hat{U_f} h_j^{o} + b_f) \textrm{,} \\
u_t^j &= \sigma(W_u x_j^{i} + \hat{W_u} x_j^{o} + U_u h_j^{i} + \hat{U_u} h_j^{o} + b_u) \textrm{,} \\
c_t^j &= f_t^j \odot c_{t-1}^j + i_t^j \odot u_t^j \textrm{,} \\
h_t^j &= o_t^j \odot \tanh (c_t^j) \textrm{,} \\
\end{split}
\end{equation*}
where $i_t^j$, $o_t^j$ and $f_t^j$ are the input, output and forget gates mentioned earlier. $W_x$, $\hat{W}_x$, $U_x$, $\hat{U}_x$, $b_x$, where $x \in \{i, o, f, u\}$, are model parameters.

\subsection{Recurrent steps}

Using the above state transition mechanism, information from each node propagates to all its neighboring nodes after each step. 
Therefore, for the worst case where the input graph is a chain of nodes, the maximum number of steps necessary for information from one arbitrary node to reach another is equal to the size of the graph. 
We experiment with different transition steps to study the effectiveness of global encoding. 

Note that unlike the sequence LSTM encoder, our graph encoder allows parallelization in node-state updates, and thus can be highly efficient using a GPU\@.
It is general and can be potentially applied to other tasks, including sequences, syntactic trees and cyclic structures. 

\subsection{Input Representation}
\label{sec:input}

Different from sequences, the edges of an AMR graph contain labels, which represent relations between the nodes they connect, and are thus important for modeling the graphs.
Similar with Section \ref{eq:base_inp_2}, we adopt two different ways for calculating the representation for each edge $(i,j,l)$: 
\begin{align}
x_{i,j}^l &= W_4 \Big( [e_l; e_i] \Big) + b_4 \\
x_{i,j}^l &= W_4 \Big( [e_l; e_i; h_i^c] \Big) + b_4 \textrm{,}
\end{align}
where $e_l$ and $e_i$ are the embeddings of edge label $l$ and source node $v_i$, $h_i^c$ denotes the last hidden state of the character LSTM over $v_i$, and $W_4$ and $b_4$ are trainable parameters.
The equations correspond to Equations \ref{eq:base_inp} and \ref{eq:base_inp_2} in Section \ref{sec:base_inp}, respectively.

\subsection{Decoder}


We adopt the attention-based LSTM decoder as described in Section \ref{sec:base_dec}.
Since our graph encoder generates a sequence of graph states, only the last graph state is adopted in the decoder.
In particular, we make the following changes to the decoder. First, each attention vector becomes $a_j=[h_T^j; x_j]$, where $h_T^j$ is the last state for node $v_j$. 
Second, the decoder initial state $s_{-1}$ is the average of the last states of all nodes.

\subsection{Integrating the copy mechanism}
\label{sec:copy}

Open-class tokens, such as dates, numbers and named entities, account for a large portion in the AMR corpus. 
Most appear only a few times, resulting in a data sparsity problem.
To address this issue, \newcite{konstas-EtAl:2017:Long} adopt anonymization for dealing with the data sparsity problem.
In particular, they first replace the subgraphs that represent dates, numbers and named entities (such as ``(q / quantity :quant 3)'' and ``(p / person :name (n / name :op1 ``Ryan''))'') with predefined placeholders (such as ``num\_0'' and ``person\_name\_0'') before decoding, and then recover the corresponding surface tokens (such as ``3'' and ``Ryan'') after decoding.
This method involves hand-crafted rules, which can be costly.

\subparagraph{Copy}
We find that most of the open-class tokens in a graph also appear in the corresponding sentence, and thus adopt the copy mechanism \cite{gulcehre-EtAl:2016:P16-1,gu-EtAl:2016:P16-1} to solve this problem.
The mechanism works on top of an attention-based RNN decoder by integrating the attention distribution into the final vocabulary distribution.
The final probability distribution is defined as the interpolation between two probability distributions:
\begin{equation}
P_{final} = \theta_t P_{vocab} + (1-\theta_t) P_{attn}\textrm{,}
\end{equation}
where $\theta_t$ is a switch for controlling generating a word from the vocabulary or directly copying it from the input graph.
$P_{vocab}$ is the probability distribution of directly generating the word, as defined in Equation \ref{eq:pvocab}, and 
$P_{attn}$ is calculated based on the attention distribution $\alpha_t$ by summing the probabilities of the graph nodes that contain identical concept.
Intuitively, $\theta_t$ is relevant to the current decoder input $e_{t}$ and state $s_t$, and the context vector $\mu_t$.
Therefore, we define it as:
\begin{equation}
\theta_t = \sigma(w_\mu^T \mu_t + w_s^T s_t + w_e^T e_{t} + b_5)\textrm{,}
\end{equation}
where vectors $w_\mu$, $w_s$, $w_e$ and scalar $b_{5}$ are model parameters.
The copy mechanism favors generating words that appear in the input.
For AMR-to-text generation, it facilitates the generation of dates, numbers, and named entities that appear in AMR graphs.

\subparagraph{Copying vs anonymization}
Both copying and anonymization alleviate the data sparsity problem by handling the open-class tokens.
However, the copy mechanism has the following advantages over anonymization:
(1) anonymization requires significant manual work to define the placeholders and heuristic rules both from subgraphs to placeholders and from placeholders to the surface tokens, 
(2) the copy mechanism automatically learns what to copy, while anonymization relies on hard rules to cover all types of the open-class tokens, 
and (3) the copy mechanism is easier to adapt to new domains and languages than anonymization.

\section{Training and decoding}

We train our models using the cross-entropy loss over each gold-standard output sequence $W^*=w_1^*, \dots, w_t^*, \dots, w_M^*$:
\begin{equation}
l = -\sum_{t=1}^M \log p(w_t^*|w_{t-1}^*,\dots,w_1^*,X;\theta)\textrm{,}
\end{equation}
where $X$ is the input graph, and $\theta$ is the model parameters.
Adam \cite{kingma2014adam} with a learning rate of 0.001 is used as the optimizer, and the model that yields the best devset performance is selected to evaluate on the test set.
Dropout with rate 0.1 is used during training.
Beam search with beam size to 5 is used for decoding.
Both training and decoding use Tesla K80 GPUs.

\section{Experiments}

\subsection{Data}

We use a standard AMR corpus (LDC2015E86) as our experimental dataset, which contains 16,833 instances for training, 1368 for development and 1371 for test. Each instance contains a sentence and an AMR graph.

Following \newcite{konstas-EtAl:2017:Long}, we supplement the gold data with large-scale automatic data.
We take Gigaword as the external data to sample raw sentences, and train our model on both the sampled data and LDC2015E86.
We adopt \newcite{konstas-EtAl:2017:Long}'s strategy for sampling sentences from Gigaword, and choose JAMR \cite{flanigan-EtAl:2016:SemEval} to parse selected sentences into AMRs, as the AMR parser of \newcite{konstas-EtAl:2017:Long} only works on the anonymized data.
For training on both sampled data and LDC2015E86, we also follow the method of \newcite{konstas-EtAl:2017:Long}, which is fine-tuning the model on the AMR corpus after every epoch of pretraining on the gigaword data.

\subsection{Settings}

We extract a vocabulary from the training set, which is shared by both the encoder and the decoder.
The word embeddings are initialized from Glove pretrained word embeddings \cite{pennington2014glove} on Common Crawl, and are not updated during training.
Following existing work, we evaluate the results with the BLEU metric \cite{papineni2002bleu}.

For model hyperparameters, we set the graph state transition number as 9 according to development experiments.
Each node takes information from at most 10 neighbors. 
The hidden vector sizes for both encoder and decoder are set to 300 (They are set to 600 for experiments using large-scale automatic data).
Both character embeddings and hidden layer sizes for character LSTMs are set 100, and at most 20 characters are taken for each graph node or linearized token.

\subsection{Development experiments}
\label{sec:comp_sys}

\begin{table}
\centering
\begin{tabular}{l|c|c}
\hline
Model & BLEU & Time \\
\hline
\hline
Seq2seq & 18.8 & 35.4s \\ 
Seq2seq+copy & 19.9 & 37.4s \\ 
Seq2seq+charLSTM+copy & 20.6 & 39.7s \\ 
\hline
Graph2seq & 20.4 & 11.2s \\ 
Graph2seq+copy & 22.2 & 11.1s \\ 
Graph2seq+Anon & 22.1 & 9.2s \\ 
Graph2seq+charLSTM+copy & \textbf{22.8} & 16.3s \\ 
\hline
\end{tabular}
\caption{\textsc{Dev} BLEU scores and decoding times.}
\label{tab:dev_res}
\end{table}

As shown in Table \ref{tab:dev_res}, we compare our model with a set of baselines on the AMR devset to demonstrate how the graph encoder and the copy mechanism can be useful when training instances are not sufficient.
\emph{Seq2seq} is the sequence-to-sequence baseline described in Section \ref{sec:base}.
\emph{Seq2seq+copy} extends \emph{Seq2seq} with the copy mechanism, 
and \emph{Seq2seq+charLSTM+copy} further extends \emph{Seq2seq+copy} with character LSTM\@.
\emph{Graph2seq} is our graph-to-sequence model, \emph{Graph2seq+copy} extends \emph{Graph2seq} with the copy mechanism, and \emph{Graph2seq+charLSTM+copy} further extends \emph{Graph2seq+copy} with the character LSTM\@.
We also try \emph{Graph2seq+Anon}, which applies our graph-to-sequence model on the anonymized data from \newcite{konstas-EtAl:2017:Long}.

\subparagraph{The graph encoder}
As can be seen from Table \ref{tab:dev_res}, the performance of \emph{Graph2seq} is 1.6 BLEU points higher than \emph{Seq2seq}, which shows that our graph encoder is effective when applied alone.
Adding the copy mechanism (\emph{Graph2seq+copy} vs \emph{Seq2seq+copy}), the gap becomes 2.3.
This shows that the graph encoder learns better node representations compared to the sequence encoder, which allows attention and copying to
function better.

Applying the graph encoder together with the copy mechanism gives a gain of 3.4 BLEU points over the baseline (\emph{Graph2seq+copy} vs \emph{Seq2seq}).
The graph encoder is consistently better than the sequence encoder no matter whether character LSTMs are used.

We also list the encoding part of decoding times on the devset, 
as the decoders of the seq2seq and the graph2seq models are similar,
so the time differences reflect efficiencies of the encoders.
Our graph encoder gives consistently better efficiency compared with the sequence encoder, showing the advantage of parallelization.

\subparagraph{The copy mechanism}
Table \ref{tab:dev_res} shows that the copy mechanism is effective on both the graph-to-sequence and the sequence-to-sequence models.
Anonymization gives comparable overall performance gains on our graph-to-sequence model as the copy mechanism (comparing \emph{Graph2seq+Anon} with \emph{Graph2seq+copy}).
However, the copy mechanism has several advantages over anonymization as discussed in Section \ref{sec:copy}.

\subparagraph{Character LSTM}
Character LSTM helps to increase the performances of both systems by roughly 0.6 BLEU points.
This is largely because it further alleviates the data sparsity problem by handling unseen words, which may share common substrings with in-vocabulary words.

\subsection{Effectiveness on graph state transitions}

We report a set of development experiments for understanding the graph LSTM encoder.

\begin{figure}[t]
\centering
\includegraphics[width=0.95\linewidth]{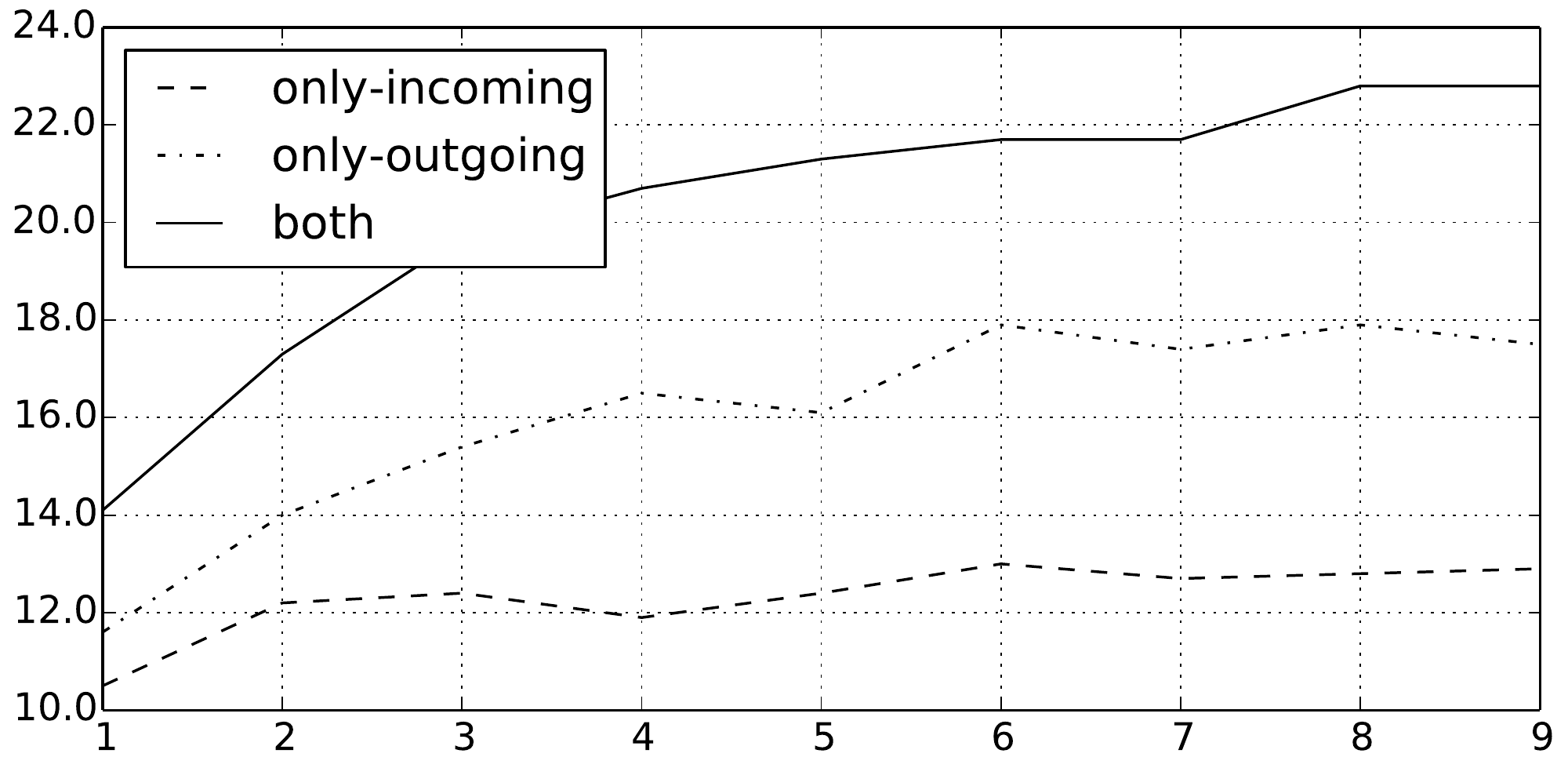}
\vspace{-1.0em}
\caption{\textsc{Dev} BLEU scores against transition steps for the graph encoder.}
\label{fig:iters}
\end{figure}

\begin{figure}[t]
\centering
\includegraphics[width=0.95\linewidth]{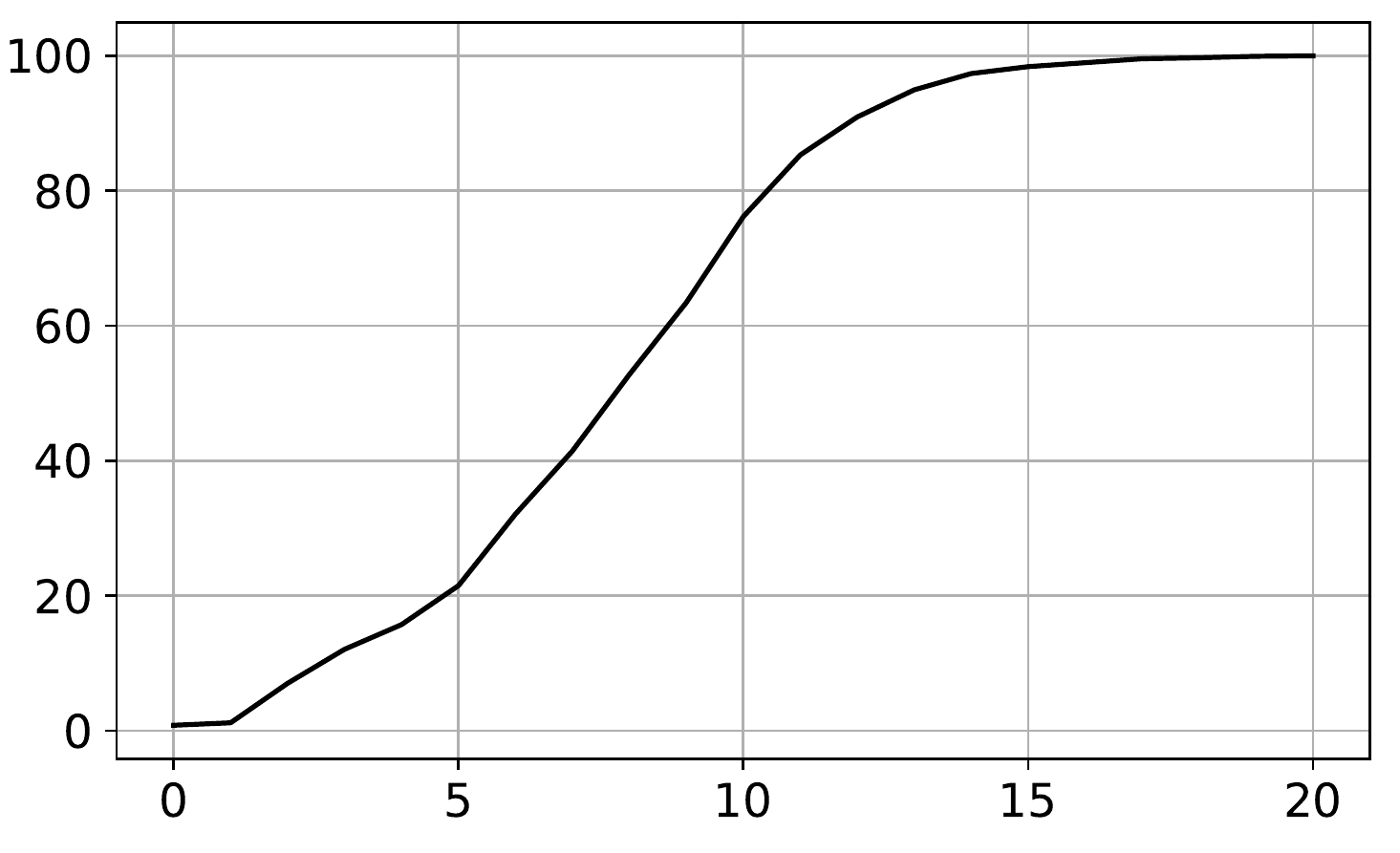}
\vspace{-1.0em}
\caption{Percentage of \textsc{Dev} AMRs with different diameters.}
\label{fig:depth}
\vspace{-1.0em}
\end{figure}

\subparagraph{Number of iterations}
We analyze the influence of the number of state transitions to the model performance on the devset.
Figure \ref{fig:iters} shows the BLEU scores of different state transition numbers, when both incoming and outgoing edges are taken for calculating the next state (as shown in Figure \ref{fig:encoder}).
The system is \emph{Graph2seq+charLSTM+copy}.
Executing only 1 iteration results in a poor BLEU score of 14.1.
In this case the state for each node only contains information about immediately adjacent nodes.
The performance goes up dramatically to 21.5 when increasing the iteration number to 5.
In this case, the state for each node contains information of all nodes within a distance of 5.
The performance further goes up to 22.8 when increasing the iteration number from 5 to 9,
where all nodes with a distance of less than 10 are incorporated in the state for each node.

\subparagraph{Graph diameter}
We analyze the percentage of the AMR graphs in the devset with different graph diameters and show the cumulative distribution in Figure \ref{fig:depth}.
The diameter of an AMR graph is defined as the longest distance between two AMR nodes.\footnote{The diameter of single-node graphs is 0.}
Even though the diameters for less than 80\% of the AMR graphs are less or equal than 10, our development experiments show that it is not necessary to incorporate the whole-graph information for each node.
Further increasing state transition number may lead to additional improvement. 
We do not perform exhaustive search for finding the optimal state transition number.

\subparagraph{Incoming and outgoing edges}
As shown in Figure \ref{fig:iters}, we analyze the efficiency of state transition when only incoming or outgoing edges are used.
From the results, we can see that there is a huge drop when state transition is performed only with incoming or outgoing edges.
Using edges of one direction, the node states only contain information of ancestors or descendants.
On the other hand, node states contain information of ancestors, descendants, and siblings if edges of both directions are used.
From the results, we can conclude that not only the ancestors and descendants, but also the siblings are important for modeling the AMR graphs.
This is similar to observations on syntactic parsing tasks \cite{mcdonald-crammer-pereira:2005:ACL}, where sibling features are adopted.

We perform a similar experiment for the \emph{Seq2seq+copy} baseline by only executing single-directional LSTM for the encoder.
We observe BLEU scores of 11.8 and 12.7 using only forward or backward LSTM, respectively.
This is consistent with our graph model in that execution using only one direction leads to a huge performance drop. 
The contrast is also reminiscent of using the normal input versus the reversed input in neural machine translation \citep{sutskever2014sequence}.

\subsection{Results}

\begin{table}
\centering
\begin{tabular}{l|c}
\hline
Model & BLEU \\
\hline
\hline
PBMT  & 26.9 \\
SNRG  & 25.6 \\
Tree2Str  & 23.0 \\
MSeq2seq+Anon  & 22.0 \\
Graph2seq+copy & 22.7 \\
Graph2seq+charLSTM+copy & 23.3 \\
\hline
MSeq2seq+Anon (200K) & 27.4 \\
MSeq2seq+Anon (2M)  & 32.3 \\
MSeq2seq+Anon (20M)  & \textbf{33.8} \\
\hline
Seq2seq+charLSTM+copy (200K) & 27.4 \\
Seq2seq+charLSTM+copy (2M) & 31.7 \\
Graph2seq+charLSTM+copy (200K) & 28.2 \\
Graph2seq+charLSTM+copy (2M) & \textbf{33.6}\tablefootnote{It was 33.0 at submission, and has been improved.} \\
\hline
\end{tabular}
\caption{\textsc{Test} results. ``(200K)'', ``(2M)'' and ``(20M)'' represent training with the corresponding number of additional sentences from Gigaword.}
\label{tab:global_res}
\end{table}

Table \ref{tab:global_res} compares our final results with existing work.
\emph{MSeq2seq+Anon} \cite{konstas-EtAl:2017:Long} is an attentional multi-layer sequence-to-sequence model trained with the anonymized data.
\emph{PBMT} \cite{pourdamghani-knight-hermjakob:2016:INLG} adopts a phrase-based model for machine translation \cite{koehn2003statistical} on the input of linearized AMR graph, \emph{SNRG} \cite{song-EtAl:2017:Short} uses synchronous node replacement grammar for parsing the AMR graph while generating the text, and \emph{Tree2Str} \cite{jeff2016amrgen} converts AMR graphs into trees by splitting the re-entrances before using a tree transducer to generate the results.

\emph{Graph2seq+charLSTM+copy} achieves a BLEU score of 23.3, which is 1.3 points better than \emph{MSeq2seq+Anon} trained on the same AMR corpus.
In addition, our model without character LSTM is still 0.7 BLEU points higher than \emph{MSeq2seq+Anon}.
Note that \emph{MSeq2seq+Anon} relies on anonymization, which requires additional manual work for defining mapping rules, thus limiting its usability on other languages and domains.
The neural models tend to underperform statistical models when trained on limited (16K) gold data, but performs better with scaled silver data \cite{konstas-EtAl:2017:Long}.

Following \newcite{konstas-EtAl:2017:Long}, we also evaluate our model using both the AMR corpus and sampled sentences from Gigaword.
Using additional 200K or 2M gigaword sentences, \emph{Graph2seq+charLSTM+copy} achieves BLEU scores of 28.2 and 33.0, respectively, which are 0.8 and 0.7 BLEU points better than \emph{MSeq2seq+Anon} using the same amount of data, respectively.
The BLEU scores are 5.3 and 10.1 points better than the result when it is only trained with the AMR corpus, respectively.
This shows that our model can benefit from scaled data with automatically generated AMR graphs, and it is more effective than \emph{MSeq2seq+Anon} using the same amount of data.
Using 2M gigaword data, our model is better than all existing methods.
\newcite{konstas-EtAl:2017:Long} also experimented with 20M external data, obtaining a BLEU of 33.8.
We did not try this setting due to hardware limitations.
The \emph{Seq2seq+charLSTM+copy} baseline trained on the large-scale data is close to \emph{MSeq2seq+Anon} using the same amount of training data, yet is much worse than our model.

\subsection{Case study}

We conduct case studies for better understanding the model performances.
Table \ref{tab:examples} shows example outputs of sequence-to-sequence  (\emph{S2S}), graph-to-sequence (\emph{G2S}) and graph-to-sequence with copy mechanism (\emph{G2S+CP}). 
\emph{Ref} denotes the reference output sentence, and \emph{Lin} shows the serialization results of input AMRs.
The best hyperparameter configuration is chosen for each model.

For the first example, \emph{S2S} fails to recognize the concept ``a / account'' as a noun and loses the concept ``o / old'' (both are underlined).
The fact that ``a / account'' is a noun is implied by ``a~/~account :mod (o~/~old)'' in the original AMR graph.
Though directly connected in the original graph, their distance in the serialization result (the input of \emph{S2S}) is 26, which 
may be why \emph{S2S} makes these mistakes.
In contrast, \emph{G2S} handles ``a~/~account'' and ``o~/~old'' correctly.
In addition, the copy mechanism helps to copy ``look-over'' from the input, which rarely appears in the training set.
In this case, \emph{G2S+CP} is incorrect only on hyphens and literal reference to ``anti-japanese war'', although the meaning is fully understandable.

For the second case, both \emph{G2S} and \emph{G2S+CP} correctly generate the noun ``agreement'' for ``a~/ agree'' in the input AMR, while \emph{S2S} fails to.
The fact that ``a~/~agree'' represents a noun can be determined by the original graph segment ``p / provide :ARG0 (a / agree)'', which indicates that ``a / agree'' is the subject of ``p / provide''.
In the serialization output, the two nodes are close to each other.
Nevertheless, \emph{S2S} still failed to capture this structural relation, which reflects the fact that a sequence encoder is not designed to explicitly model hierarchical information encoded in the serialized graph.
In the training instances, serialized nodes that are close to each other can originate from neighboring graph nodes, or distant graph nodes, which prevents the decoder from confidently deciding the correct relation between them.
In contrast, \emph{G2S} sends the node ``p / provide'' simultaneously with relation ``ARG0'' when calculating hidden states for ``a / agree'', which facilitates the yielding of ``the agreement provides''.

\begin{table}[t!] \small
\centering
\begin{tabularx}{0.5\textwidth}{X}
\hline
(p / possible-01 :polarity - \\
~~~~:ARG1 (l / look-over-06 \\
~~~~~~~~:ARG0 (w / we) \\
~~~~~~~~:ARG1 (a / \underline{account}-01 \\
~~~~~~~~~~~~:ARG1 (w2 / war-01 \\
~~~~~~~~~~~~~~~~:ARG1 (c2 / country :wiki ``Japan'' \\
~~~~~~~~~~~~~~~~~~~~:name (n2 / name :op1 ``Japan'')) \\
~~~~~~~~~~~~~~~~:time (p2 / previous) \\
~~~~~~~~~~~~~~~~:ARG1-of (c / call-01 \\
~~~~~~~~~~~~~~~~~~~~:mod (s / so))) \\
~~~~~~~~~~~~:mod (o / \underline{old})))) \\
\textbf{Lin}: possible :polarity - :arg1 ( look-over :arg0 we :arg1 ( \underline{account} :arg1 ( war :arg1 ( country :wiki japan :name ( name :op1 japan ) ) :time previous :arg1-of ( call :mod so ) ) :mod \underline{old} ) ) \\
\textbf{Ref}: we can n't look over the old accounts of the previous so-called anti-japanese war . \\
\textbf{S2S}: we can n't be able to account the past drawn out of japan 's entire war .\\
\textbf{G2S}: we can n't be able to do old accounts of the previous and so called japan war.\\
\textbf{G2S+CP}: we can n't look-over the old accounts of the previous so called war on japan . \\
\hline
(p / provide-01 \\
~~~~:ARG0 (a / \underline{agree}-01) \\
~~~~:ARG1 (a2 / and \\
~~~~~~~~:op1 (s / staff \\
~~~~~~~~~~~~:prep-for (c / center \\
~~~~~~~~~~~~~~~~:mod (r / research-01))) \\
~~~~~~~~:op2 (f / fund-01 \\
~~~~~~~~~~~~:prep-for c))) \\
\textbf{Lin}: provide :arg0 \underline{agree} :arg1 ( and :op1 ( staff :prep-for ( center :mod research ) ) :op2 ( fund :prep-for center ) ) \\
\textbf{Ref}: the agreement will provide staff and funding for the research center .\\
\textbf{S2S}: agreed to provide research and institutes in the center .\\
\textbf{G2S}: the agreement provides the staff of research centers and funding . \\
\textbf{G2S+CP}: the agreement provides the staff of the research center and the funding .\\
\hline
\end{tabularx}
\caption{Example system outputs.}
\label{tab:examples}
\end{table}

\section{Related work}

Among early statistical methods for AMR-to-text generation, \newcite{jeff2016amrgen} convert input graphs to trees by splitting re-entrances, and then translate the trees into sentences 
with a tree-to-string transducer.
\newcite{song-EtAl:2017:Short} use a synchronous node replacement grammar to parse input AMRs and generate sentences at the same time.
\newcite{pourdamghani-knight-hermjakob:2016:INLG} linearize input graphs by breadth-first traversal, and then use a phrase-based machine translation system\footnote{http://www.statmt.org/moses/} to generate results by translating linearized sequences.

Prior work using graph neural networks for NLP include the use graph convolutional networks (GCN) \cite{kipf2017semi} for semantic role labeling \cite{marcheggiani-titov:2017:EMNLP2017}, neural machine translation \cite{bastings-EtAl:2017:EMNLP2017} and graph-to-sequence learning \cite{xu2018graph2seq}. 
Both GCN and the graph LSTM update node states by exchanging information between neighboring nodes within each iteration. 
However, our graph state LSTM adopts gated operations for making updates, while GCN uses a linear transformation.
Intuitively, the former has better learning power than the later.
Another major difference is that our graph state LSTM keeps a cell vector for each node to remember all history.
The contrast between our model with GCN is reminiscent of the contrast between RNN and CNN. We leave empirical comparison of their effectiveness to future work. 
In this work our main goal is to show that graph LSTM encoding of AMR is superior compared with sequence LSTM.

Closest to our work, \newcite{TACL1028} modeled syntactic and discourse structures using DAG LSTM, which can be viewed as extensions to tree LSTMs \cite{tai-socher-manning:2015:ACL-IJCNLP}.
The state update follows the sentence order for each node, and has sequential nature.
Our state update is in parallel.
In addition, \newcite{TACL1028} split input graphs into separate DAGs before their method can be used. 
To our knowledge, we are the first to apply an LSTM structure to encode AMR graphs.

The recurrent information exchange mechanism in our state transition process is remotely related to the idea of loopy belief propagation (LBP) \cite{murphy1999loopy}. 
However, there are two major differences. 
First, messages between LSTM states are gated neural node values, rather than probabilities in LBP\@. 
Second, while the goal of LBP is to estimate marginal probabilities, the goal of information exchange between graph states in our LSTM is to find neural representation features, which are directly optimized by a task objective. 

In addition to NMT \cite{gulcehre-EtAl:2016:P16-1}, the copy mechanism has been shown effective on tasks such as dialogue \cite{gu-EtAl:2016:P16-1}, summarization \cite{see-liu-manning:2017:Long} and question generation \citep{song-naacl-18}.
We investigate the copy mechanism on AMR-to-text generation.

\section{Conclusion}

We introduced a novel graph-to-sequence model for AMR-to-text generation.
Compared to sequence-to-sequence models, which require linearization of AMR before decoding, a graph LSTM is leveraged to directly model full AMR structure.
Allowing high parallelization, the graph encoder is more efficient than the sequence encoder.
In our experiments, the graph model outperforms a strong sequence-to-sequence model, achieving the best performance.

\paragraph{Acknowledgments}
We thank the anonymized reviewers for the insightful comments, and the Center for Integrated Research Computing (CIRC) of University of Rochester for special reservations of computation resources.

\bibliography{naaclhlt2018}
\bibliographystyle{acl_natbib}

\end{document}